\documentclass[letterpaper]{article} 
\usepackage{aaai2026}  
\nocopyright
\usepackage{times}  
\usepackage{helvet}  
\usepackage{courier}  
\usepackage[hyphens]{url}  
\usepackage{graphicx} 
\usepackage{natbib}  
\usepackage{caption} 
\usepackage{algorithm}
\usepackage{algorithmic}
\usepackage{xcolor}
\usepackage{alltt} 
\usepackage{amsmath}
\usepackage{amssymb}
\usepackage{dsfont}
\usepackage{multirow}
\usepackage{booktabs}
\usepackage{newfloat}
\usepackage{listings}
\DeclareCaptionStyle{ruled}{labelfont=normalfont,labelsep=colon,strut=off} 
\floatstyle{ruled}
\newfloat{listing}{tb}{lst}{}
\floatname{listing}{Listing}
\title{An Entity Linking Agent for Question Answering}

\author {
    Yajie Luo\textsuperscript{\rm 1}\thanks{Equal contribution.},
    Yihong Wu\textsuperscript{\rm 1}\footnotemark[1],
    Muzhi Li\textsuperscript{\rm 2},
    Jia Ao Sun\textsuperscript{\rm 1},
    Xinyu Wang\textsuperscript{\rm 3},
    Liheng Ma\textsuperscript{\rm 3, 4},
    Yingxue Zhang\textsuperscript{\rm 5},
    Jian-Yun Nie\textsuperscript{\rm 1}
 }
 
\affiliations {
    \textsuperscript{\rm 1}Université de Montréal\\
    \textsuperscript{\rm 2}The Chinese University of Hong Kong\\
    \textsuperscript{\rm 3}McGill University\\
    \textsuperscript{\rm 4}Mila - Quebec AI Institute\\
    \textsuperscript{\rm 5}Huawei Noah’s Ark Lab\\
    \{yajie.luo, yihong.wu, jia.ao.sun, jian-yun.nie\}@umontreal.ca,
    mzli@cse.cuhk.edu.hk,\\  
    \{xinyu.wang5, liheng.ma \}@mail.mcgill.ca,
    yingxue.zhang@huawei.com

}

\usepackage{bibentry}

\begin{document}

\maketitle

\begin{abstract}
Some Question Answering (QA) systems rely on knowledge bases (KBs) to provide accurate answers. Entity Linking (EL) plays a critical role in linking natural language mentions to KB entries. However, most existing EL methods are designed for long contexts and do not perform well on short, ambiguous user questions in QA tasks. We propose an entity linking agent for QA, based on a Large Language Model that simulates human cognitive workflows. The agent actively identifies entity mentions, retrieves candidate entities, and makes decision. To verify the effectiveness of our agent, we conduct two experiments: tool-based entity linking and QA task evaluation. The results confirm the robustness and effectiveness of our agent.
\end{abstract}


\section{Introduction}
Most Question Answering (QA) systems follow the retrieve-then-answer paradigm, where they retrieve information from a Knowledge Base (KB) to answer a given question~\cite{mavi2024multi}.
Common KBs include textual corpora and knowledge graphs~\cite{hogan2021knowledge}.
In addition to well-established, public, general KBs such as  Wikipedia~\cite{vrandevcic2014wikidata}, Wikidata~\cite{vrandevcic2014wikidata} and Freebase~\cite{freebase}, a recent trend involves constructing KBs from local, private, and domain-specific knowledge~\cite{edge2024local}. 
Consequently, interfacing natural language with KBs is essential.

Entity Linking (EL) is the task of connecting natural language to KBs by mapping textual mentions to their corresponding entities within a KB~\cite{mihalcea2007wikify}.
Traditionally, EL systems consist of two main components: Mention Detection (MD) and Entity Disambiguation (ED)~\cite{zhang2021entqa}. MD, also known as Named Entity Recognition (NER)~\cite{al2020named}, identifies potential entities from text. Subsequently, ED links these mentions to their unique entries in the KB~\cite{pershina2015personalized}.

Early approaches to EL generally follow a ``detect-then-disambiguate'' pipeline, first employing off-the-shelf MD systems before focusing on the ED task~\cite{hoffart2011robust, van2020rel}.
They impractically assumed all detected mentions need to be addressed, not considering the error propagation.
Moreover, this separation ignores the latent dependencies between the two sub-tasks, leading to sub-optimal performance.
To address this limitation, researchers have shifted towards joint methods.
One line of research involves end-to-end systems~\cite{kolitsas2018end, fang2019joint, de2020autoregressive}.
Another approach reformulates EL as a QA problem, leveraging the retriever-reader paradigm~\cite{chen2017reading}.
In this framework, the retriever performs a role analogous to MD by proposing potential entity candidates, while the reader executes ED by selecting the correct entity from those candidates~\cite{wu2019scalable,zhang2021entqa,orlando2024relik, xiao2023instructed}.

However, the previously mentioned methods are designed for EL in long-form documents. In contrast, this work focuses on EL for QA~\cite{li2020efficient, shavarani-sarkar-2025-entity}, which requires linking entities within the concise text of a user's query.
The challenges in this QA setting are distinct; queries are typically short, ambiguous, and lack the rich context.
This necessitates reasoning over implicit type constraints~\cite{Li_Yang_Xu_Song_Jiang_Guo_Leung_King_2025} and background knowledge not explicitly provided in the question. Therefore, these unique constraints demand a novel EL method tailored for QA.

To address these challenges, we propose a LLM-based (Large Language Model) entity linking agent, inspired by recent advances in LLM agents~\cite{liu2025advances, wu2025reinforcing}.
Our approach models the human cognitive workflow for the EL task, where a person identifies potential entities, uses search tools (e.g., Wikidata, Google Knowledge Graph\footnote{https://developers.google.com/knowledge-graph} provide entity search engines) to find candidates, and then disambiguates them based on the available context.
Our EL agent is designed to replicate this process.
To achieve the target, the system must learn to \textbf{plan} by actively selecting entity mentions, \textbf{use tools} by querying a search engine, and \textbf{make decisions} by selecting the correct entity from the search results.
These capabilities align with the definition of autonomous agents~\cite{liu2025advances}.

Our approach offers three primary advantages over existing methodologies.
First, our agent-based workflow provides superior retrieval flexibility.
Prevailing methods~\cite{wu2019scalable, zhang2021entqa} almost exclusively use dense retrieval~\cite{karpukhin2020dense}.
While effective, dense retrieval struggles with certain types of entity references~\cite{sciavolino2021simple}.
In contrast, our framework is retriever-agnostic.
Specifically, we prefer lexical retrievers, e.g., BM25~\cite{bm25}, as their keyword-matching mechanism is particularly effective given the entity-centric nature of our task.
Second, by employing an LLM as its core, our agent leverages reasoning and common sense to overcome the challenge of limited context.
This capability is critical for disambiguation when explicit clues are minimal.
We further leverage Chain-of-Thought (CoT) prompting~\cite{wei2022chain} for an robust and interpretable agent decision-making process.
Furthermore, our method is highly adaptable.
It can be implemented on large-scale LLMs (e.g., GPT-4, Claude, Gemini) through few-shot learning~\cite{brown2020language} or fine-tuned on smaller models via self-play training~\cite{zelikman2022star, wu2025reinforcing}.
Our method's effectiveness is confirmed through extensive experimentation.
\section{Related Works}
\paragraph{Entity Linking} The task of entity linking is typically divided into two sequential sub-tasks: mention detection and entity disambiguation~\cite{kolitsas2018end}.
Early works in entity linking adopted graphed-based disambiguation for consistency~\cite{kulkarni2009collective, yosef2011aida, ganea2016probabilistic}, typically by constructing a graph over candidate entities from all mentions and performing joint inference to ensure coherent linking.
With the advent of deep learning, research has shifted toward retrieval-based EL frameworks.
These models encode both mention contexts and entity candidates into a shared embedding space, enabling efficient similarity-based retrieval~\cite{sil2018neural, gillick2019learning, le2019boosting, ganea2017deep, wu2019scalable}. 
Specifically, ELQ~\cite{li2020efficient} adopted a bi-encoder architecture to jointly perform mention detection and linking in one pass.
RefinED~\cite{ayoola2022refined} is an end-to-end entity linking model that jointly performs mention detection, fine-grained entity typing, and disambiguation in a single forward pass. 
Entqa~\cite{zhang2021entqa} introduced a retrieval-and-reading framework inspired by QA, where a dense retriever first selects candidate entities, followed by a reader that verifies each entity in context. 
In contrast to prior methods that rely on dense retrieval, our approach is retriever-agnostic and leverages a LLM as an agent to identify mentions, utilize external tools, and make decisions.

\paragraph{Question Answering}
Open-domain question answering systems have traditionally employed a retrieve-and-read pipeline, first retrieving context from a large corpus before generating an answer~\cite{chen2017reading}.
Early research in QA was largely divided into two distinct tasks: information retrieval, which focuses on finding documents that support an answer, and Machine Reading Comprehension (MRC), which involves extracting an answer from a given passage~\cite{zeng2020survey}. 
The emergence of LLMs~\cite{chowdhery2023palm, achiam2023gpt, touvron2023llama}, has reduced the need for explicit MRC modules in QA tasks, due to their strong reasoning and language understanding abilities.
Nonetheless, LLMs often struggle with obscure or long-tail questions, necessitating the use of external retrieval systems for augmentation~\cite{lewis2020retrieval}.
However, conventional retrieval methods have persistent weaknesses. Sparse retrievers cannot fully grasp semantic meaning or support fuzzy matching~\cite{karpukhin2020dense}, while dense retrievers can be ineffective for entity-centric questions~\cite{sciavolino2021simple}.
To address this gap, recent research has explored entity retrieval as an alternative approach~\cite{shavarani2024entity, salnikov2023large, mohammed2018strong, lukovnikov2019pretrained}.
Following this line of work, our research investigates the application of entity linking for QA.

\paragraph{LLM Agent}
With the rapid progress of LLMs, recent research has introduced autonomous LLM agents, which can independently perceive goals, plan actions, interact with external tools, and make decisions without step-by-step instruction~\cite{yao2023react, hong2023metagpt}. More examples include self-supervised tool-usage learning via API calls~\cite{schick2023toolformer}, a debate-driven multi-agent framework~\cite{du2023improving}, self-organizing agents for open-ended tasks~\cite{chen2024s}, role-based collaborative planning agents~\cite{hao2025chatllm}, all of them demonstrate the effectiveness of autonomous LLM agents in addressing complex, real-world tasks.
Inspired by these advances, we design an autonomous LLM-based agent tailored for entity linking in question answering,  capable of identifying entity mentions, retrieving relevant candidates, and making context-aware linking decisions.

\section{Methodology}
\label{methodology}
\begin{figure*}[htbp]
    \centering
    \includegraphics[width=\textwidth, height=0.45\textheight, keepaspectratio]{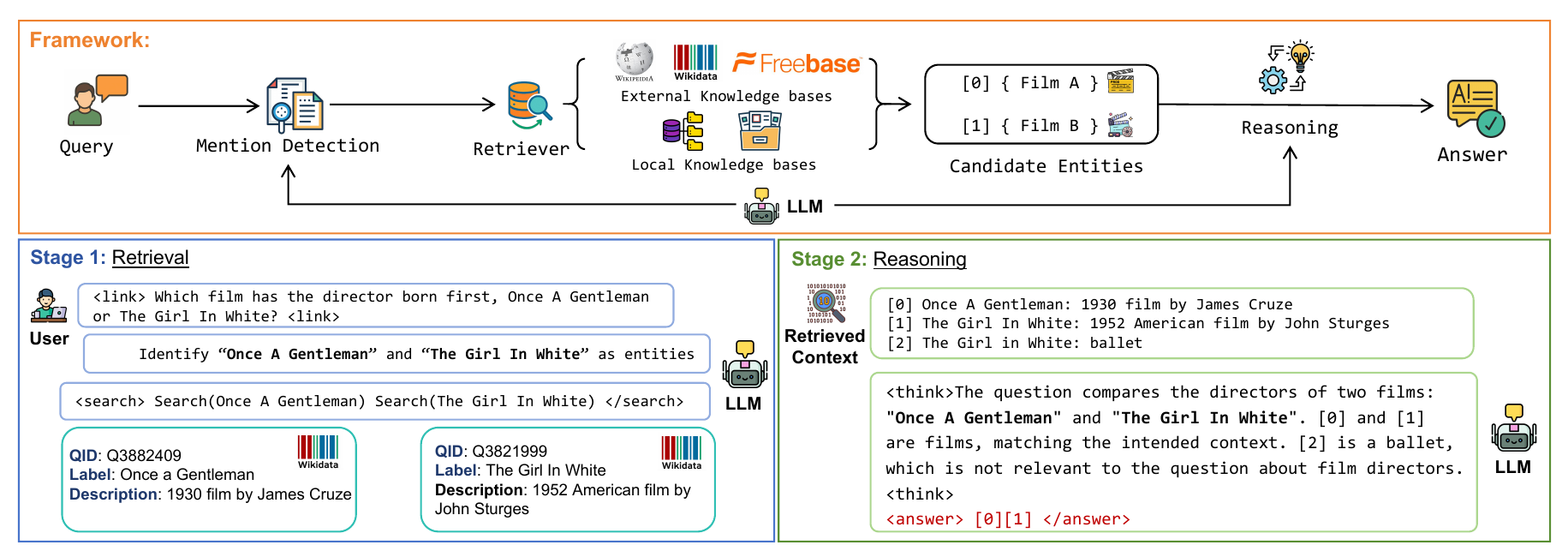} 
    \caption{Entity Linking Agent (ELA) Framework.} 
    \label{fig:framework}
\end{figure*}
\subsection{Task Definition}
A knowledge base~\cite{lan2021survey} is a repository of information, such as knowledge graphs~\cite{hogan2021knowledge} or a text corpus.
An entity is a unique object within a KB.
Let $\mathcal{T}$ denote the text space.
Given a document $d \in \mathcal{T}$ and a KB $\mathcal{K}$, the task of \textbf{entity linking}~\cite{shen2015entity} is to identify entities $e_i \in \mathcal{K}$ mentioned in $d$.
This task is typically decomposed into two sub-tasks: mention detection and entity disambiguation.
MD identifies text spans, or potential mentions $m_i \in \mathcal{T}$, within document $d$ that may refer to an entity.
ED maps each detected mention $m_i$ to its corresponding entity $e_i$ in the knowledge base $\mathcal{K}$.

For this work, we assume $\mathcal{K}$ is sufficiently comprehensive to contain all relevant entities.
For mentions that do not have a corresponding entity in the KB (i.e., false mentions), we allow the target entity an empty entity~\cite{zhang2021entqa}.

\subsection{Entity Linking Agents}
This work focuses on EL within the context of question answering, who aims to develop systems providing direct answers to user queries~\cite{choi2018quac}.
In contrast to traditional EL, which targets long-form, context-rich documents, EL for QA concentrate on short, ambiguous queries.
In this scenario, the objective is not to identify every possible entity, but rather to pinpoint the entities of interest, who are crucial for resolving the query.
The challenge is maintaining high accuracy despite the limited contextual information.
This necessitates new EL methods with advanced language understanding capabilities.

To address this challenge, we propose \textbf{ELA} (Entity Linking Agent), an LLM-based agent for entity linking.
This agent leverages the inherent language understanding and reasoning capabilities of LLMs to identify entity mentions and select contextually relevant candidates.
As illustrated in Figure~\ref{fig:framework}, our agent's workflow mimics the human approach to this task.
Given a query $q\in \mathcal{T}$, the agent first identifies potential entity mentions $\{m_i\}$.
It then utilizes a search tool \texttt{Search()} to retrieve a Top-K list of candidate entities for each mention.
Subsequently, the agent select the most appropriate entity from each list based on the query $q$ and the other retrieved candidates.
This workflow aligns with the retriever-reader paradigm, which is widely adopted in prior works~\cite{wu2019scalable, zhang2021entqa}.
The following sections detail each component of this agent.

\paragraph{Retrieval}
The retrieval stage comprises two tasks: mention recognition and tool use.
Given a query $q$, the agent needs to identify a set of potential mentions, $\{m_i\}$, within the text.
Each mention is defined as a span of text within $q$.
For example, in the query, \textit{"Which film has the director born first, Once A Gentleman or The Girl In White?"}, the agent should identify \textit{"Once A Gentleman"} and \textit{"The Girl In White"} as mentions, as these named entities are critical for answering the question.
For the tool-use task, the agent is instructed to use the detected mentions as inputs for a search engine.
To achieve this, the LLM is prompted to generate its output in a structured format, \texttt{Search(X)}, where \texttt{X} is the mention to be searched.
This format allows for straightforward parsing using regular expressions.
For efficiency, multiple mentions are processed in a single forward pass of the LLM.
The specific instructions and a complete example are provided in Appendix.

A key feature of our workflow is that it is retriever-agnostic.
While prior works often relied on dense retrievers~\cite{wu2019scalable, zhang2021entqa}, such models can exhibit limitations in precise entity recognition—a critical function in EL~\cite{sciavolino2021simple}—despite their strength in capturing semantic meaning.
Our design, therefore, is not bound to a specific retriever, which offers significant flexibility.
The framework can accommodate any search backend, from public APIs like Wikidata or the Google Knowledge Graph to specialized retrievers.
The optimal choice depends on the nature of the mentions.
For explicit mentions that act as strong lexical signals, sparse retrievers like BM25~\cite{bm25} may be preferred. Conversely, when entities are mentioned implicitly, dense retrievers are likely to be more effective due to their semantic capabilities.

\paragraph{Reader}

In the reader stage, the agent selects relevant entities $e_i$ from a Top-K candidate set $\{c_j\}_{j=1}^K$ for each mention, based on the query $q$ and retrieved context. Each candidate $c_j$ comprises a title and a short description (e.g., \textit{"The girl in white: 1952 American film by John Sturges"}).
The agent evaluates these candidates and identifies semantically relevant matches, necessitating an understanding of both the query intent and entity descriptions.

Rather than directly prompting the LLM for answers, we employ CoT reasoning~\cite{wei2022chain}.
Prior work~\cite{guo2025deepseek, jaech2024openai} demonstrates that CoT enhances model reasoning and problem-solving capabilities.
In our framework, the CoT rationale resembles: \textit{"The query concerns the director of two films... While both [x] and [y] reference 'the girl in white,' [x] corresponds to a Van Gogh painting..."}
Additionally, CoT provides partial explainability, aiding error diagnosis and agent decision analysis. The agent’s CoT outputs are encapsulated within the \texttt{<think>} tag.

\paragraph{Implementation}
Despite LLMs' versatility, we still need to instruct them to follow the predefined workflow.
In-context learning~\cite{brown2020language} is a way to let LLM follow the instruction without training.
We prepare a few examples and inject them via text prompts so that LLMs know how to response in different cases.
This method is ad-hoc, easy to implement; however, it might suffer in some cases like that a small size models cannot completely understand the instructions or that those few-shot prompts seriously slow the inference speed.

To mitigate these issues, we introduce a lightweight RL-style self-play fine-tuning method~\cite{zelikman2022star}. First, we generate trajectories (query, CoT, answer) via in-context learning, then filter for those with exact matches to gold entities. These high-quality trajectories train the LLM using cross-entropy loss~\cite{radford2018improving}, enabling smaller models to achieve competitive performance. Post fine-tuning, few-shot examples can be omitted during inference, eliminating their computational overhead without sacrificing accuracy.

LLM-based agents are often criticized for computational inefficiency, but recent advances in AI infrastructure alleviate this concern.
Leveraging optimizations like vLLM~\cite{kwon2023efficient}, our system processes 1,000 queries in under 5 minutes on a single \texttt{NVIDIA H100} GPU for a 7B model.
Further improvement is achievable with multi-GPU deployment or next-generation hardware.

\subsection{Entity Linking Meets Question Answering}
Traditional information retrieval is dominated by two paradigms: sparse and dense retrieval, both of which have inherent limitations.
Sparse methods, such as BM25~\cite{bm25}, operate on lexical matching and thus struggle with semantic nuances, failing to retrieve relevant documents that do not share exact keywords with the query.
Conversely, dense retrievers~\cite{karpukhin2020dense}, which rely on semantic similarity, can be ineffective for entity-centric questions where precise identification is critical~\cite{sciavolino2021simple}.

To address these gaps, we consider EL as a complementary retrieval mechanism.
The key advantage of EL is its ability to ground entity mentions in a query to their canonical entries in a knowledge base.
Unlike methods that match text, EL can precisely pinpoint a specific entity, a property that is highly advantageous for downstream tasks like knowledge graph QA~\cite{lan2021survey} and factoid QA~\cite{stelmakh2022asqa}.
Accordingly, we integrate our ELA agent into a complete QA pipeline to verify its effectiveness as a high-precision retrieval component.

This work also raises a fundamental question: should EL be treated as an independent, task-agnostic process, or should its objectives be defined by the downstream application? For QA, we contend that the goal of EL is to identify all "entities of interest" within a query.
However, the definition of an "entity of interest" is itself ambiguous.

Consider the question: \textit{When did Michael Jordan return to the NBA?} A traditional NER approach would identify both \textit{Michael Jordan} and \textit{NBA} as valid entities to be linked.
From a QA perspective, however, only \textit{Michael Jordan} is the core topic entity.
Retrieving general information about the NBA is unlikely to yield the specific answer, making it a noisy entity in this context.

This distinction is critical as it directly impacts the evaluation of EL systems for QA. Most established EL benchmarks~\cite{hoffart2011robust, hoffart2012kore} are designed for document-level linking, where the objective is to identify every entity in a long text. A key pitfall of this approach is its failure to consider the utility for a downstream task. We argue that a successful EL system for QA must not only link entities but also distinguish between those essential for answering the question and those that are peripheral noise.

The development of such systems, however, is hindered by a scarcity of carefully constructed, QA-centric benchmarks. We therefore call for the creation of more sophisticated evaluation frameworks that measure the practical contribution of EL to downstream QA performance.

\section{Experiments}
To evaluate the effectiveness of our entity linking agent, we conduct two experiments.
The first experiment examines the agent’s ability to leverage existing search tools (Tool Use) for identifying and linking entities mentioned in the input question.
The second experiment applies the entity linking technique to a real-world QA task.

\begin{table*}[t]
\centering
\small
\setlength{\tabcolsep}{5pt}  
\renewcommand{\arraystretch}{1.2} 

\begin{tabular}{clccc ccc ccc}
\toprule
\multirow{2}{*}{Source} & \multirow{2}{*}{Methods}
    & \multicolumn{3}{c}{2Wiki}
    & \multicolumn{3}{c}{WebQSP}
    & \multicolumn{3}{c}{CWQ} \\
\cmidrule(lr){3-5} \cmidrule(lr){6-8} \cmidrule(lr){9-11}
 &  & Precision & Recall & Acc
    & Precision & Recall & Acc
    & Precision & Recall & Acc \\
\midrule

 & Azure     & \underline{92.49} & \underline{91.00} & \underline{88.20} & 11.10 & 10.75 & 10.30 & 78.10 & 79.07 & 60.60 \\
 Wikipedia& ELQ \cite{li2020efficient}       & 75.05 & 74.35 & 72.1 & \underline{91.05} & \underline{92.05} & \underline{86.60} & 81.77 & 82.35 & 68.40 \\
 & ReFinED \cite{ayoola2022refined}   & 76.63 & 79.40 & 69.10 & 89.93 & 90.80 & 84.90 & \underline{81.46} & \underline{83.48} & \underline{68.90} \\
\midrule

 & \textbf{ELA (Ours)}             &      &       &      &      &      &      &      &      &      \\
 & w/ Llama-3.1-8B (few-shot)         & 77.71 & 68.30 & 55.20 & 46.40 & 45.75 & 42.60 & 50.04 & 44.12 & 35.50 \\
 & w/ Llama-3.1-8B (fine-tuned)       & 84.45 & 81.40 & 76.60 & 83.49 & 81.78 & 78.25 & 78.00 & 68.02 & 55.80 \\
Wikidata & w/ Llama-3.3-70B           & 86.65 & 88.05 & 78.70 & 79.12 & 83.83 & 74.22 & 73.39 & 69.64 & 59.81 \\
 & w/ DeepSeek-V3                     & 88.03 & 88.05 & 79.70 & 86.40 & 88.20 & 83.40 & 81.44 & 82.72 & 68.40 \\
 & w/ GPT-4.1                         & 83.52 & 82.25 & 75.80 & 80.14 & 81.15 & 78.20 & 65.83 & 63.35 & 56.70 \\
\midrule

 & \textbf{ELA (Ours)}             &      &       &      &      &      &      &      &      &      \\
 & w/ Llama-3.1-8B (few-shot)         & 67.17 & 57.05 & 43.40 & 50.20 & 49.35 & 47.00 & 46.49 & 43.18 & 38.84 \\
 & w/ Llama-3.1-8B (fine-tuned)       & 89.65 & 89.35 & 86.50 & 84.52 & 82.65 & 79.20 & 78.65 & 67.75 & 54.50 \\
Wikipedia & w/ Llama-3.3-70B         & 85.70 & 85.50 & 80.70 & 74.77 & 77.77 & 69.66 & 51.46 & 50.34 & 44.72 \\
 & w/ DeepSeek-V3                     & 91.92 & 91.40 & 88.00 & \textbf{91.18} & \textbf{92.55} & \textbf{88.20} &\textbf{82.43}  & \textbf{84.04} & \textbf{73.05} \\
 & w/ GPT-4.1                         & \textbf{97.06} & \textbf{96.45} & \textbf{93.72} & 84.45 & 84.70 & 82.40 & 69.30 & 65.85 & 59.30 \\
\bottomrule
\end{tabular}
\caption{Entity Linking Agent performance. 
\textit{ELA} stands for \textit{Entity Linking Agent}. 
\textit{Bolded values} indicate the best performance for each evaluation metric.
\underline{Underlined values} highlight the second-best performance.
Acc denotes accuracy.
We omit the percent sign ($\%$) from all results.
\textit{w/} denotes the base model used with the agent (e.g., \textit{w/ GPT-4.1} refers to using GPT-4.1 as the backbone language model). }
\label{tab:entity-linking}
\end{table*}

\subsection{Tool Use}
Unlike previous works~\cite{zhang2021entqa, wu2019scalable} that are tightly coupled with a specific dense retriever, our agentic method is retriever-agnostic. This design provides the flexibility to operate with any existing search tool. To demonstrate this capability, we perform entity linking using the search engines of both Wikidata~\cite{vrandevcic2014wikidata} and Wikipedia.

\paragraph{Datasets}
We evaluate our method on three widely-used question answering (QA) benchmarks: 2WikiMultiHopQA (2Wiki)~\cite{ho2020constructing}, WebQSP~\cite{webqsp}, and ComplexWebQuestions (CWQ)~\cite{cwq}. 2Wiki is a multi-hop QA benchmark built using Wikidata and Wikipedia. WebQSP is a single-hop dataset, while CWQ is its multi-hop extension featuring complex questions that require multi-step reasoning. In these datasets, each instance consists of a natural language question and its corresponding topic entities, which the entity linking system must identify.
A challenge with WebQSP and CWQ is their reliance on Freebase~\cite{bollacker2008freebase}, a knowledge graph that has been deprecated. Therefore, following prior work~\cite{li2020efficient}, we converted their original Freebase entity IDs to Wikidata QIDs using a mapping derived from the Freebase dump\footnote{https://developers.google.com/freebase}. In addition, to satisfy our requirement, we further preprocessed the datasets. Details can be found in Appendix.

\paragraph{Metrics}
We use precision (Prec), recall, and accuracy (Acc) to assess the performance of our agent.
Given the set of gold entities $\mathit{g}$ and the set of predicted entities $\mathit{p}$, these metrics are defined as the follows:
\begin{equation*}
     \text{Prec} = \frac{|\mathit{p} \cap \mathit{g}|}{|\mathit{p}|}  \;,\; \text{Recall} = \frac{|\mathit{p} \cap \mathit{g}|}{|\mathit{g}|} \;,\; \text{Acc} = \mathds{1}(\mathit{p} == \mathit{g}) \;,
\end{equation*}
where $\mathds{1}$ is an indicator function ($1$ for exact match between $\mathit{g}$ and $\mathit{p}$ $0$ for otherwise).
Precision measures how many predictions are correct.
Recall measures how many true entities are covered by the prediction.
Accuracy, a.k.a. Micro F1, is the exact match between gold and predicted entities.


\paragraph{Baselines}
To evaluate the effectiveness of our approach, we compare it against three representative entity linking (EL) baselines: the commercial Azure Entity Linking system\footnote{https://docs.azure.cn/en-us/ai-services/language-service/entity-linking/overview}, ELQ~\cite{li2020efficient}, and ReFinED~\cite{ayoola2022refined}.
\begin{itemize}
    \item The Azure Entity Linking system is a closed-source commercial service from Microsoft that links entity mentions to Wikipedia.
    \item ELQ is an end-to-end model optimized for questions, employing a bi-encoder architecture to jointly perform mention detection and linking.
    \item ReFinED is a high-performance linker that combines dense retrieval with a transformer-based model for efficient and accurate entity disambiguation.
\end{itemize}
We chose ELQ and ReFinED because they are specifically designed for or evaluated on QA datasets, making them more suitable than the numerous EL methods developed for document-level tasks.
The Azure system was included as it has been widely adopted in prior QA research~\cite{ma2025thinkongraph}.
We used the official code and  re-implement both ELQ and ReFinED in our experiments.

\begin{table*}[ht]
\centering
\renewcommand{\arraystretch}{1.2}
\setlength{\tabcolsep}{8pt}
\begin{tabular}{lccc ccc}
\toprule
\multirow{2}{*}{\textbf{Methods}}       
& \multicolumn{3}{c}{\textbf{TriviaQA}}      
& \multicolumn{3}{c}{\textbf{PopQA}}         \\
\cmidrule(lr){2-4} \cmidrule(lr){5-7}
& \textbf{Hit@1} & \textbf{EM} & \textbf{F1} 
& \textbf{Hit@1} & \textbf{EM} & \textbf{F1} \\
\midrule
Naive Gen, CoT                     & --    & 77.60 & 84.20 & --    & 34.00 & 43.76 \\
BM25                               & 55.88 & 75.09 & 80.61 & 62.00 & 42.10 & 50.06 \\ 
\multicolumn{7}{l}{Dense Retrieval} \\
\quad w/ intfloat/e5-large-v2   & 42.03 & 72.37 & 77.53 & 72.20 & 49.10 & 58.99 \\
\quad w/ BAAI/bge-base-en-v1.5    & 67.39 & 78.37 & 83.85 & 75.60 & 51.70 & 61.09 \\
\quad w/ BAAI/bge-large-en-v1.5   & 69.42 & 81.23 & 86.55 & 77.20 & 52.10 & 62.06 \\
\textbf{Entity Linking (Ours)}     & 64.60 & 79.90 & 85.93 & 76.50 & 53.40 & 64.05 \\
\bottomrule
\end{tabular}
\caption{Performance of entity linking in QA.
\textit{Naive Gen} denotes naive generation with Chain-of-Thought prompting. 
We omit the percent sign ($\%$) from all results.
\textit{w/} indicates the use of different dense retrievers.}
\label{tab:el_performance}
\end{table*}
\paragraph{Implementation Details}
We experiment with four LLMs: Llama-3.1-8B-Instruct (Llama-8B), Llama-3.3-70B-Instruct (Llama-70B), DeepSeek-V3, and GPT-4.1. Among them, Llama-8B is used for both few-shot prompting and fine-tuning. While the others are evaluated under few-shot setting with chain-of-thought prompting.
To accelerate inference, we use vLLM~\cite{kwon2023efficient}. We set the temperature to $0.7$, top-p to $0.8$, and repetition penalty 1.05 if applicable.
For retrieval, we set $k$ to $50$ for the Top-$k$ returned list.
To enhance the performance of smaller model, we perform full-parameter self-play fine-tuning on the Llama-8B model.
We randomly selected 3000 samples from 2Wiki training set. For each instance, we first generate (query, chain of thought, answer) trajectories via context learning and retain only 1500 examples whose predicted entities exactly matched the gold entities for fine-tuning.
More training details are in Appendix.

\paragraph{Results} Table~\ref{tab:entity-linking} presents the precision, recall, and Accuracy metrics for four methods across three datasets.
Our results consistently show that our proposed ELA method outperforms all three baselines-Azure, ELQ, and RefinED on the three benchmarks.
Specially, on the 2wiki dataset, for example, ELA with GPT-4.1 model achieves 93.72\% accuracy, significantly higher than baselines.
Even on the more challenging CWQ dataset, our methods remain competitive, with DeepSeek-V3 achieving 73.05\% accuracy, was about 4.15\% higher than the next best ReFinED method. These results validate the effectiveness of our ELA framework for the entity linking task.
Moreover, large language models—such as LLaMA-3.3-70B, DeepSeek-V3, and GPT-4.1—consistently outperform smaller models like LLaMA-3.1-8B under the same few-shot prompting setting, suggesting that larger models are better at following complex entity linking instructions. 
We also observed that the best and second-best performance are consistently achieved when Wikipedia is used as knowledge source. This may be attributed to the fact that Wikidata may provide many candidate entities for a given mention, introducing more noise into the disambiguation process. Wikipedia, by contrast, aligns more naturally with QA datasets.
However, this difference from the inherent characteristics of the knowledge bases themselves, rather than from our proposed method.

On the WebQSP dataset, the Azure Entity linking systems perform extremely poor with the accuracy of 10.3\%, despite strong performance on the other datasets.
This is likely because WebQSP is entirely lowercased, while Azure system is known to be case-sensitive. In constrast, ELA maintains high performance on WebQSP, highlighting the robustness of our approach.


\paragraph{In-Context Learning}
We initially explored using In-Context Learning (ICL), a method where LLMs learn new tasks from a few examples without parameter updates, to guide our entity linking agent.
However, our experiments with an 8B model revealed several significant challenges with this approach.
First, the model frequently failed to follow our instruction, causing errors in the downstream processing pipeline.
Second, it struggled with fine-grained semantic distinctions, confusing closely related but distinct entities.
Third, the model exhibited a tendency to over-infer or "hallucinate" queries.
For instance, after identifying a named film, it would proceed to query for related but unmentioned entities like the film's director.
To overcome these limitations, we considered two alternatives.
Using a larger model, such as GPT-4, could improve instruction following but would incur prohibitive inference costs.
Another one is to manually refine prompts and examples that better align with the specific requirements of our task through prompt engineering.
But it requires significant time and manual effort.
So we propose to finetune a local model specifically for EL tasks.

\paragraph{Fine-tuning}

We fine-tuned a LLaMA-3 8B model using a small, high-quality dataset of only 1,500 trajectories sampled from the 2Wiki training set.
The resulting model exhibits three key advantages: high performance, strong cross-dataset generalization, and remarkable data efficiency.
First, despite its relatively small size, our model achieves state-of-the-art performance.
Notably, on WebQSP dataset, it performs better than the much larger LLaMA-3.3-70B model (74.22\%) and even slightly outperforms GPT-4.1 (78.20\%) with an accuracy of 78.25\% in wikidata source. In addition to accuracy, the 8B model achieves much faster inference than larger models.
Second, the model shows strong generalization capabilities. Although fine-tuned exclusively on 2Wiki, it achieves competitive results on both the WebQSP and CWQ datasets.
Finally, this performance is achieved with exceptional data efficiency.
The success with only 1,500 training examples underscores that our method does not require large-scale, costly fine-tuning.
While not the primary focus of this work, we believe that scaling the training data and further optimizing the training process could yield even greater performance, which we leave for future work.

\subsection{Entity Linking in Question Answering}
\label{sec_el4qa}
In this experiment, we apply EL as a retrieval method in QA task to evaluate whether it can effectively assist the task.
\paragraph{Datasets}
To verify the effectiveness of EL in real QA tasks, we further conduct experiments on two widely-used datasets: TriviaQA~\cite{joshi2017triviaqa} and PopQA~\cite{mallen2022not}. 
TriviaQA is a large scale open-domain question answering dataset collected from trivia websites. 
PopQA is a popularity-aware open-domain-QA benchmark with each answer grounded in Wikipedia.
For TriviaQA, we directly use the provided evidence documents as the retrieval corpus. In contrast, PopQA does not include associated context documents, making it unsuitable for direct use as a retrieval corpus. To address this, we construct a corpus for PopQA by crawling the Wikipedia pages corresponding to the entities linked to each question.

\paragraph{Metrics}
We evaluate the performance of applying entity linking to QA task using Hit@1, Exact Match (EM), and F1 score.
Hit@1 is a retrieval metric that measures whether the top-1 predicted document matches the gold document. Since entity linking returns only a single document, we report Hit@1 for fair comparison with other retrieval-based methods.
EM and F1 score are standard metrics in QA~\cite{rajpurkar2016squad}. 
EM measures the exact string match between the predicted and gold answers.
The F1 score balances precision and recall, measuring the overlap between prediction and gold answers.

\paragraph{Baselines}
To assess the effectiveness of our proposed approach, we compare it against three types of QA system: a naive generation method with chain-of-thought prompting~\cite{wei2022chain} by asking LLMs the question directly, sparse retrievers with LLMs, and dense retrievers with LLMs.
We use the DeepSeek-V3 as the base model to generate final answers based on the retrieved documents in all settings.
In the naive generation baseline, LLM answers questions only based on its own knowledge.
Next, we adopt the BM25 algorithm~\cite{robertson2009probabilistic} as a sparse baseline to retrieve relevant documents from Wikipedia.
For dense retrieval, we evaluate three embedding models: intfloat/e5-large-v2~\cite{wang2022text}, BAAI/bge-base-en-v1.5\footnote{https://huggingface.co/BAAI/bge-base-en-v1.5}~\cite{bge_m3}, and BAAI/bge-large-en-v1.5\footnote{https://huggingface.co/BAAI/bge-large-en-v1.5}~\cite{bge_m3}.
All models are used to encode full Wikipedia page texts into dense vector representations.
To improve retrieval efficiency, we adopt Scalable Vector Search (SVS) library~\cite{aguerrebere2023similarity} for approximate nearest neighbor search. 

\paragraph{Implementation Details}
For this experiment, we use DeepSeek-V3 as the base model for our proposed ELA method. Unlike the previous setup, which used existing search tools, we now employ a custom-built entity search system.
Our search system is built on a BM25 retriever that leverages the title-content structure of the Wikipedia corpus. We indexed the unique title of every Wikipedia article, enabling the retriever to perform a lexical match between a query (i.e., an entity mention) and the article titles. For each query, the system returns a ranked list of the top k=35 candidate entities, where each candidate consists of the article's title and its first paragraph as a concise description. Once our ELA agent identifies the correct entity from this list, the full text of the corresponding article is retrieved and provided as context to the LLM to generate the final answer.

\paragraph{Results}
Table~\ref{tab:el_performance} presents the performance comparison of our method against several baselines on the TriviaQA and PopQA datasets. Our approach outperforms the naive chain-of-thought generation baseline and BM25 on the PopQA dataset, and also exceeds all dense retrieval methods except for BAAI/bge-large-en-v1.5 model. On TriviaQA dataset, our method slightly behind the dense retrievers in Hit@1, it still achieves competitive results, with 64.60\% Hit@1 and 79.90\% EM.
It is important to note that our primary research objective is not to achieve state-of-the-art QA performance, but rather to investigate whether EL can serve as an effective retrieval mechanism to access essential information required for answering questions.  
To keep the setup simple, we adopt a basic strategy using a lightweight BM25 retriever to fetch the top-k Wikipedia titles relevant to the given entity name.
Despite this simplicity, our method achieves strong Hit@1 performance, especially on PopQA. These results demonstrate the feasibility of incorporating entity linking into QA as a retrieval mechanism. Future work can explore more advanced retrieval strategies to further improve overall performance.

\section{Conclusion}
In this work, we propose a LLM-based entity linking agent that adopts a retrieval-then-reader strategy enabling it to effectively identify mentions, retrieve candidate entities, and select the most relevant entities. Our experiments show that the agent not only achieves strong performance in entity linking across multiple datasets, but also performs competitively in question answering when used as a retrieval mechanism.

\bibliography{ref}

\newpage
\clearpage
\section{Appendix}
\paragraph{Prompts for Entity Linking}
\begin{figure*}[h]
    \centering
    \includegraphics[width=\textwidth]{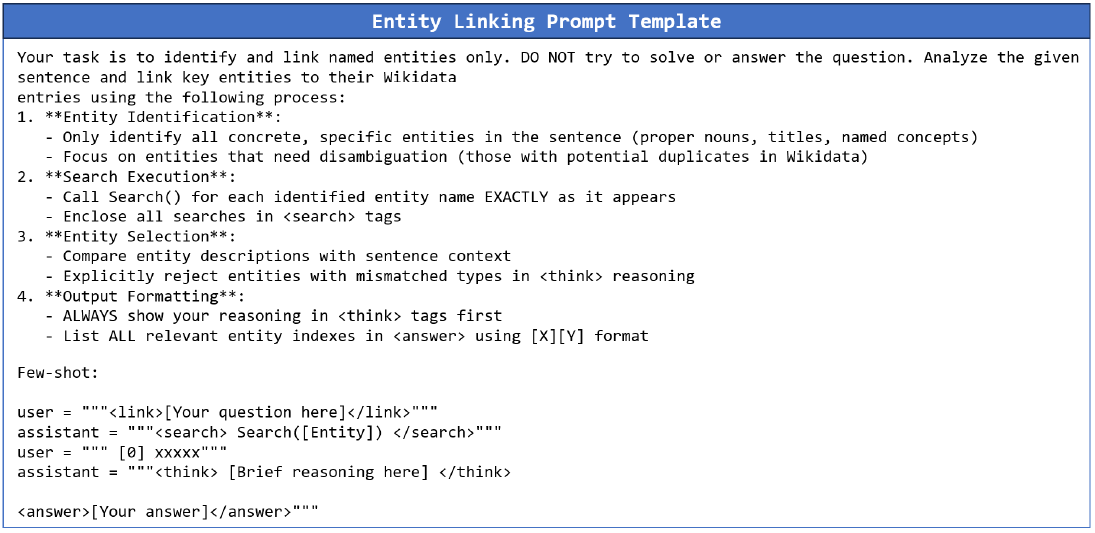}
    \caption{Entity Linking Prompts.}
    \label{fig:prompt}
\end{figure*}
The following is the complete prompt and few-shot examples used to guide the language model during entity identification and linking. It enforces strict constraints on what can be searched and linked, focusing only on explicitly mentioned named entities, as shown in Figure~\ref{fig:prompt}.

\paragraph{Dataset preprocess}
Both the WebQSP and CWQ datasets were originally constructed on top of the Freebase knowledge base. Each dataset consists of natural language questions paired with SPARQL queries. However, for the task of entity linking, explicit topic entities are required for each question, which are not provided in the original data. To obtain them, we first extract all Freebase entity IDs (MIDs) from the SPARQL queries and then apply a rule-based filtering strategy to identify the topic entity.
For both datasets, the extracted Freebase MIDs are converted to Wikidata QIDs using a mapping derived from the Freebase data dump. We will publicly release the processed datasets to facilitate future research.
\paragraph{Fine-tuning}
We fine-tune the LLaMA-3.1-8B model using full parameter training to adapt it to the entity linking task. Training is conducted using the LLaMA Factory framework with the AdamW optimizer. We use a global batch size of 128, a maximum sequence length of 512, and train for 3 epochs. The learning rate is set to 2e-5 and decayed linearly with a warm-up ratio of 0.1. 

\end{document}